\documentclass{article}

\usepackage[preprint]{neurips_2022}

\usepackage{xcolor}         
\definecolor{linkColor}{rgb}{0.18,0.39,0.62}
\usepackage[utf8]{inputenc} 
\usepackage[T1]{fontenc}    
\usepackage[colorlinks=true,linkcolor=linkColor,citecolor=linkColor,filecolor=linkColor,urlcolor=linkColor]{hyperref}       
\usepackage{url}            
\usepackage{booktabs}       
\usepackage{amsfonts}       
\usepackage{nicefrac}       
\usepackage{microtype}      
\usepackage[most]{tcolorbox}
\usepackage{enumitem}

\usepackage{amsmath}
\usepackage{amssymb}
\usepackage{mathtools}
\usepackage{amsthm}
\usepackage{graphicx}
\usepackage{subfigure}
\usepackage{xspace}
\usepackage{pifont}
\usepackage{xcolor}
\usepackage{array}

\usepackage{bm}
\usepackage{ulem}

\theoremstyle{plain}

\theoremstyle{definition}

\theoremstyle{remark}

\usepackage[capitalize,noabbrev]{cleveref}
\usepackage{framed}
\usepackage{multirow}
\usepackage{svg}
\usepackage{listings}
\usepackage{wrapfig}

\usepackage{multirow}
\usepackage{fontawesome5}

\newcommand\our{\textsc{VALL-E X}}

\newcommand{\cmark}{\ding{51}\xspace}%
\newcommand{\xmark}{\ding{55}\xspace}%

\makeatletter

\newcommand*{\@rowstyle}{}

\newcommand*{\rowstyle}[1]{
  \gdef\@rowstyle{#1}%
  \@rowstyle\ignorespaces%
}

\newcolumntype{=}{
  >{\gdef\@rowstyle{}}%
}

\newcolumntype{+}{
  >{\@rowstyle}%
}

\makeatother

\title{Speak Foreign Languages with Your Own Voice: Cross-Lingual Neural Codec Language Modeling}

\author{%
 Ziqiang Zhang$^*$ \ Long Zhou\thanks{Both authors contributed equally to this work. Correspondence: \{lozhou,shujliu,fuwei\}@microsoft.com} \ \ Chengyi Wang   \ Sanyuan Chen  \ Yu Wu \ Shujie Liu  \\ \textbf{Zhuo Chen \ Yanqing Liu \ Huaming Wang \ Jinyu Li   \ Lei He \ Sheng Zhao \ Furu Wei }
\\ Microsoft
\\ \url{https://github.com/microsoft/unilm}
}

\begin{document}

\maketitle

\begin{abstract}
We propose a \textit{cross-lingual neural codec language model}, \our{}, for cross-lingual speech synthesis. 
Specifically, we extend VALL-E~\citep{wang2023neural} and train a multi-lingual conditional codec language model to predict the acoustic token sequences of the target language speech by using both the source language speech and the target language text as prompts. 
\our{} inherits strong in-context learning capabilities and can be applied for zero-shot cross-lingual text-to-speech synthesis and  zero-shot speech-to-speech translation tasks. 
Experimental results show that it can generate high-quality speech in the target language via just one speech utterance in the source language as a prompt while preserving the unseen speaker's voice, emotion, and acoustic environment. Moreover, \our{} effectively alleviates the foreign accent problems, which can be controlled by a language ID.
Audio samples are available at \url{https://aka.ms/vallex}. 
\end{abstract}

\begin{figure*}[h]
  \centering
  \includegraphics[width=12cm]{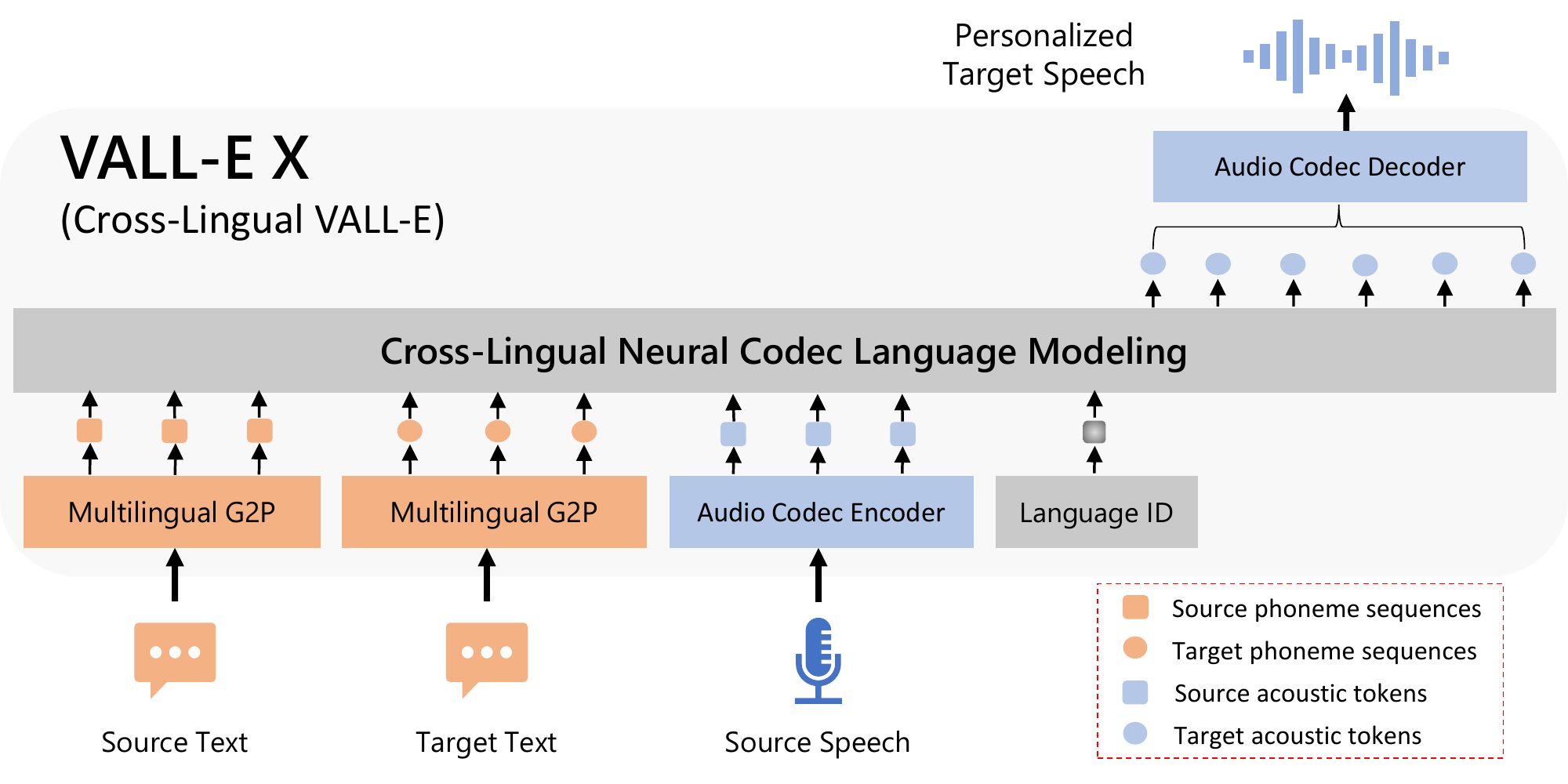}
  \caption{The overall framework of \our{}, which can synthesize personalized speech in another language for a monolingual speaker. Taking the phoneme sequences derived from the source and target text, and the source acoustic tokens derived from an audio codec model as prompts, \our{} is able to produce the acoustic tokens in the target language, which can be then decompressed to the target speech waveform. Thanks to its powerful in-context learning capabilities, \our{} does not require cross-lingual speech data of the same speakers for training, and can perform various zero-shot cross-lingual speech generation tasks, such as cross-lingual text-to-speech synthesis and speech-to-speech translation.}  
  \label{fig_vallex_framework}
\end{figure*}

\newpage
\section{Introduction}

\setcounter{footnote}{0}

Recent years have witnessed significant advancements in end-to-end text-to-speech (TTS) synthesis, and the quality of synthesized speech is even close to human parity \citep{li2019neural, ren2019fastspeech,tan2022naturalspeech}. However, these models can only generate high-quality speech for a specific speaker in a specific language. Cross-lingual speech synthesis is a new emerging task that aims to transfer the speaker's voice from one language to another. The speech quality for cross-lingual speech synthesis, especially the speaker similarity, is far behind the monolingual TTS models due to two reasons, 1) data scarcity, as it is difficult to collect multi-lingual speech data for the same speaker, and 2) model capacity, as conventional cross-lingual TTS models are not powerful enough to transfer the speaker voice, speech background, and speaker emotion from the source language speech to the target language speech.

Previous methods to tackle these challenges typically augment end-to-end TTS models with specific subnets for speaker and language control \citep{nachmani2019unsupervised, zhang2020multilingual, yang2020towards, ellinas2022cross,cai2023cross}.
For example, based on the multi-speaker TTS model, \cite{nachmani2019unsupervised} introduce multiple encoders for each language and additional loss to keep the speaker's identity. \cite{zhang2020multilingual} employs a phonemic representation to capture cross-language information and an adversarial network to disentangle speaker identities. 
\cite{yang2020towards} incorporate speaker and language networks with speaker and language IDs as input to deal with the multi-speaker and cross-lingual problems respectively.
\cite{yang2022cross} further propose a multi-task learning method with additional tasks of speaker similarity and language identification.
Moreover, \cite{cai2023cross} investigates cross-lingual multi-speaker text-to-speech synthesis with sufficient or limited bilingual speech training data.
However, the above methods fail to effectively extend to zero-shot scenarios for synthesizing target speech from the unseen source speaker,
and often suffer from low speaker similarity and L2 (second-language, or foreign) accent problems \citep{zhang2019learning,lee2022empirical}. 

\begin{table}[h]
\begin{center}
\caption{A comparison between \our{} and previous cross-lingual TTS systems.}
\label{table_compare}
\begin{tabular}{c|c|c}
\toprule
& \bf Previous Systems  & \textbf{\our{}} \\  \hline
Intermediate representation & Mel spectrogram   &   Audio codec codes \\ \hline
Training data & $<$ 13K hours   & 70K hours   \\ \hline
Speech accent &  Foreign    &  Native  \\  \hline
Speaker similarity & Relative low   &  High  \\  \hline
In-context learning & \xmark   & \cmark  \\ \hline
Zero-shot cross-lingual TTS & \xmark   & \cmark  \\ 
\bottomrule
\end{tabular}
\label{adv}
\end{center}
\end{table}

In this work, we present a novel approach to address these issues by proposing a simple yet effective cross-lingual neural codec language model, \our{}, which leverages strong in-context learning capacities to achieve high-quality zero-shot cross-lingual speech synthesis. 
Based on the knowledge learned from large-scale multi-lingual speech data, \our{} is able to transfer the speech characteristics, including the speaker's voice, emotions, and the speech background, from the source language to the target language, and also alleviate the foreign accent problems.
More specifically, we first obtain the multi-lingual speech-transcription data from existing ASR data or pseudo-labeled speech data.
Then we convert the transcriptions to phoneme sequences with a rule-based converter (G2P tool) and the speech data to acoustic tokens with an offline neural codec encoder.
Finally, we concatenate the paired phoneme and acoustic token sequences of each language and train a multi-lingual conditional language model.
As illustrated in Figure \ref{fig_vallex_framework}, after training, \our{} can predict the acoustic tokens of the target language prompted by the phoneme sequences of both languages and the acoustic tokens of the source language.
The generated acoustic token sequence is decompressed to the target speech waveform by an offline audio codec decoder.
\our{} is trained on two large-scale multi-speaker datasets\footnote{To our knowledge, the largest publicly available speech datasets for English and Chinese.}, LibriLight \citep{Kahn2020LibriLight} and WenetSpeech \citep{Zhang2022WENETSPEECH}, containing about 60,000 hours of English audiobook speech data and 10,000+ hours of multi-domain Chinese ASR data, respectively.
The combination of LibriLight and WenetSpeech makes a large multi-lingual multi-speaker multi-domain unclean speech dataset, which significantly improves the coverage of different speakers and enhances \our{}'s generalization capacity. The comparison between \our{} and the previous cross-lingual TTS systems are listed in Table \ref{table_compare}.

We conduct experiments on two kinds of cross-lingual speech generation tasks, zero-shot cross-lingual text-to-speech synthesis (XTTS), and zero-shot speech-to-speech translation (S2ST). For cross-lingual text-to-speech synthesis, the proposed \our{} is evaluated with LibriSpeech \citep{Panayotov2015Librispeech} and EMIME \citep{wester2010emime} for English and Chinese respectively, including English TTS prompted by Chinese speakers and Chinese TTS prompted by English speakers. For zero-shot speech-to-speech translation, EMIME \citep{wester2010emime} dataset is used for the evaluation of \our{} on bidirectional Chinese$\leftrightarrow$English translation tasks, and it contains bilingual audio recordings by the same speakers.
We evaluate the proposed \our{} framework from several aspects, including speaker similarity, speech quality (ASR- WER or BLEU), speech naturalness, and human evaluation (e.g., SMOS, MOS, and CMOS).
Specifically, due to the strong in-context learning capability, \our{} achieves a higher speaker similarity score than the previous SOTA model for the unseen speaker.
By training on large-scale speech-transcription data, the proposed \our{} significantly reduces the word error rate from 8.53 to 4.07 in the cross-lingual English TTS task, obtains the substantial gain of 3.17 BLEU scores than the strong baseline in S2ST tasks, and achieves better speech naturalness.
Furthermore, the human evaluation shows that our \our{} outperforms strong baselines in terms of SMOS (4.00 vs. 3.42 in XTTS, 4.12 vs. 3.06 in S2ST), CMOS (+0.24 for ours vs. baseline in XTTS), and MOS (3.87 vs. 3.81 in S2ST).
Our contributions can be summarized as follows:
\begin{itemize}
  \item We develop a cross-lingual neural codec language model \our{} with large multi-lingual multi-speaker multi-domain unclean speech data.  \our{} is a conditional cross-lingual language model predicting the target language acoustic tokens with the source language speech and target language text as prompts.
  \item The multi-lingual in-context learning framework enables \our{} to generate cross-lingual speech maintaining the unseen speaker's voice, emotion, and speech background, prompted by only one sentence in the source language.
  \item Based on the learned cross-lingual speech modeling ability with the introduced language ID, \our{} can generate speech in a native tongue for any speaker and can significantly reduce the foreign accent problem, which is a well-known problem in cross-lingual speech synthesis tasks. 
  \item We apply \our{} to zero-shot cross-lingual text-to-speech synthesis and zero-shot speech-to-speech translation tasks. Experiments show that the proposed \our{} can beat the strong baseline in terms of speaker similarity, speech quality, translation quality, speech naturalness, and human evaluation.
\end{itemize}
We encourage readers to listen to the audio samples on our demo page: \url{https://aka.ms/vallex}.

\section{Related Work}
\label{related_work}

\paragraph{Speech/Audio Synthesis}

With the rapid development and application of neural networks, speech and audio synthesis have made tremendous progress with different network frameworks, such as WaveNet \citep{oord2016wavenet}, HiFi-GAN \citep{kong2020hifi}, and Diffwave \cite{kong2020diffwave}.
Academic and industrial communities also pay increasing attention to synthesizing speech or sound from text, namely text-to-speech (TTS) \citep{li2019neural,ren2019fastspeech} or text-to-sound \citep{yang2022diffsound,kreuk2022audiogen}.
Recently, it is emerging to apply discrete audio representation learning to audio synthesis, e.g., AudioGen \citep{kreuk2022audiogen} and AudioLM \citep{borsos2022audiolm}.
AudioGen, consisting of an audio encoder, a text encoder, a Transformer encoder, and an audio decoder, is an autoregressive audio generation model with textual descriptions as inputs.
AudioLM reviews high-quality audio generation as unidirectional language modeling. In AudioLM, the input audio is mapped to semantic tokens using w2v-BERT \citep{chung2021w2v} and acoustic tokens using SoundStream \citep{zeghidour2021soundstream}.
Through three subsequent stages, AudioLM can accomplish speech continuation, acoustic generation, unconditional generation tasks, and so on.
The most related work to ours is VALL-E \citep{wang2023neural}, which was recently proposed to utilize a neural codec language model to achieve monolingual text-to-speech synthesis. Trained on large-scale speech data, VALL-E shows a strong in-content learning capability and can synthesize high-quality personalized speech prompted by a short recording of an unseen speaker.
Different from the above work, this paper focuses on cross-lingual speech synthesis, and the goal is to retain the source language speaker's voice in the synthesized speech of the target language.

\paragraph{Cross-Lingual TTS}
In cross-lingual speech synthesis, the goal is to synthesize the speech of another language for a monolingual speaker, which is more challenging than conventional monolingual TTS \citep{nachmani2019unsupervised, zhang2020multilingual,yang2022cross,cai2023cross}.
By using shared phonemic input representation across languages and incorporating an adversarial objective to disentangle the speaker's identity and speech content, \cite{zhang2019learning} is able to achieve cross-lingual voice cloning within limited speakers.
\cite{liu2020multi} also investigate the cross-lingual speech synthesis with speakers' voices enrolled in their native language. In this system, they achieve it using a Tacotron-based synthesizer with a speaker encoder module and introduce a shared phoneme set with IPA to enhance the cross-lingual capability.
Aiming at improving the speaker similarity between the synthesized speech and the recordings of the native speaker, authors in \cite{yang2022cross} propose multi-task learning by jointly training speaker classification and cross-lingual TTS models.
\cite{cai2023cross} explores cross-lingual multi-speaker speech synthesis under the scenarios of sufficient and limited bilingual training data. In the data-limited scenario, they employ a series of modules including a linguistic feature classifier, a speaker representation extractor, a non-autoregressive multi-speaker voice conversion module, and a neural vocoder, to achieve cross-lingual synthesis.
Although previous work has made considerable achievements in cross-lingual TTS, they still suffer from the issue of low speaker similarity and the lack of zero-shot ability.
In contrast, leveraging large-scale multi-lingual multi-speaker ASR data, our proposed framework with a neural codec language model demonstrates a strong in-context learning ability to alleviate the above issues.

\paragraph{Speech to Speech Translation (S2ST)}

S2ST aims to translate the speech of one language to the speech of another language. The initial research and application mainly focus on cascaded S2ST systems \citep{lavie1997janus,nakamura2006atr,wahlster2013verbmobil}, consisting of speech recognition (ASR), machine translation (MT), and speech synthesis (TTS) models. Recently, end-to-end S2ST models have been explored \citep{jia2019direct,lee2021direct,jia2021translatotron,lee2021textless,wei2022joint,huang2022transpeech,li2022textless}, achieving the direct conversion from source speech to target speech.
However, there is still an unsolved problem to reserve the source sound characteristics (e.g. speaker, emotion, and the speech background) in generated speech.
This challenge is largely due to the zero-shot nature as the bilingual speech data from the same speakers are hard to collect.
Though researchers have put much effort into constructing speech-to-speech translation corpora, such as Voxvopule \citep{wang2021voxpopuli}, CVSS \citep{jia2022cvss}, and SpeechMatrix \citep{duquenne2022speechmatrix}, they are either synthesized from text or mined from multilingual speech corpora thus can not meet the requirement that bilingual data come from the same speakers.
At the same time, Translatotron \citep{jia2019direct} tries to synthesize target speech conditioned by the speaker embedding extracted from the source speech, but it misses richer voice information due to the limitation of the speaker embedding.
Translatotron 2 \citep{jia2021translatotron} retrains the speaker voices relying on the pseudo bilingual speech data of the same speakers generated by multi-speaker TTS systems, while the synthetic speech does not completely simulate the speech of the real world.
To address these challenges, we propose to equip the cross-lingual neural codec language model with translation modules and show its zero-shot capability to reserve the sound characteristics in the S2ST task.

\section{Cross-Lingual Codec Language Model}
\label{section_Method}

In this section, we will first present the background, namely conditional codec language model VALL-E, and then introduce the framework of \our{}, followed by the multi-lingual training and cross-lingual inference approaches.

\subsection{Background}

Our \our{} is the cross-lingual version of text-to-speech synthesizer VALL-E \citep{wang2023neural}, which was recently proposed to leverage a neural codec language model to achieve text-to-speech synthesis.
Unlike conventional TTS methods that adopt the continuous regression task, e.g., mel-spectrogram generation, VALL-E regards TTS as a conditional language modeling task with neural codec codes, i.e. acoustic tokens, as an intermediate representation of speech.
VALL-E employs two-stage modeling, which first generates the codec codes of the first quantizer of EnCodec \citep{defossez2022high} from the paired phoneme sequences using an autoregressive language model, and then generates the codes of the rest quantizers in parallel using a non-autoregressive model.
After training on the large-scale English speech-transcription dataset LibriLight, VALL-E shows strong in-context learning capabilities.
It can generate personalized speech by taking only a 3-second speech fragment as a prompt.
Based on VALL-E, our \our{} extend to train a cross-lingual neural codec language model, enabling zero-shot cross-lingual capability and supporting cross-lingual TTS or speech-to-speech translation tasks. 

\begin{figure*}[!t]
  \centering
  \includegraphics[width=14cm]{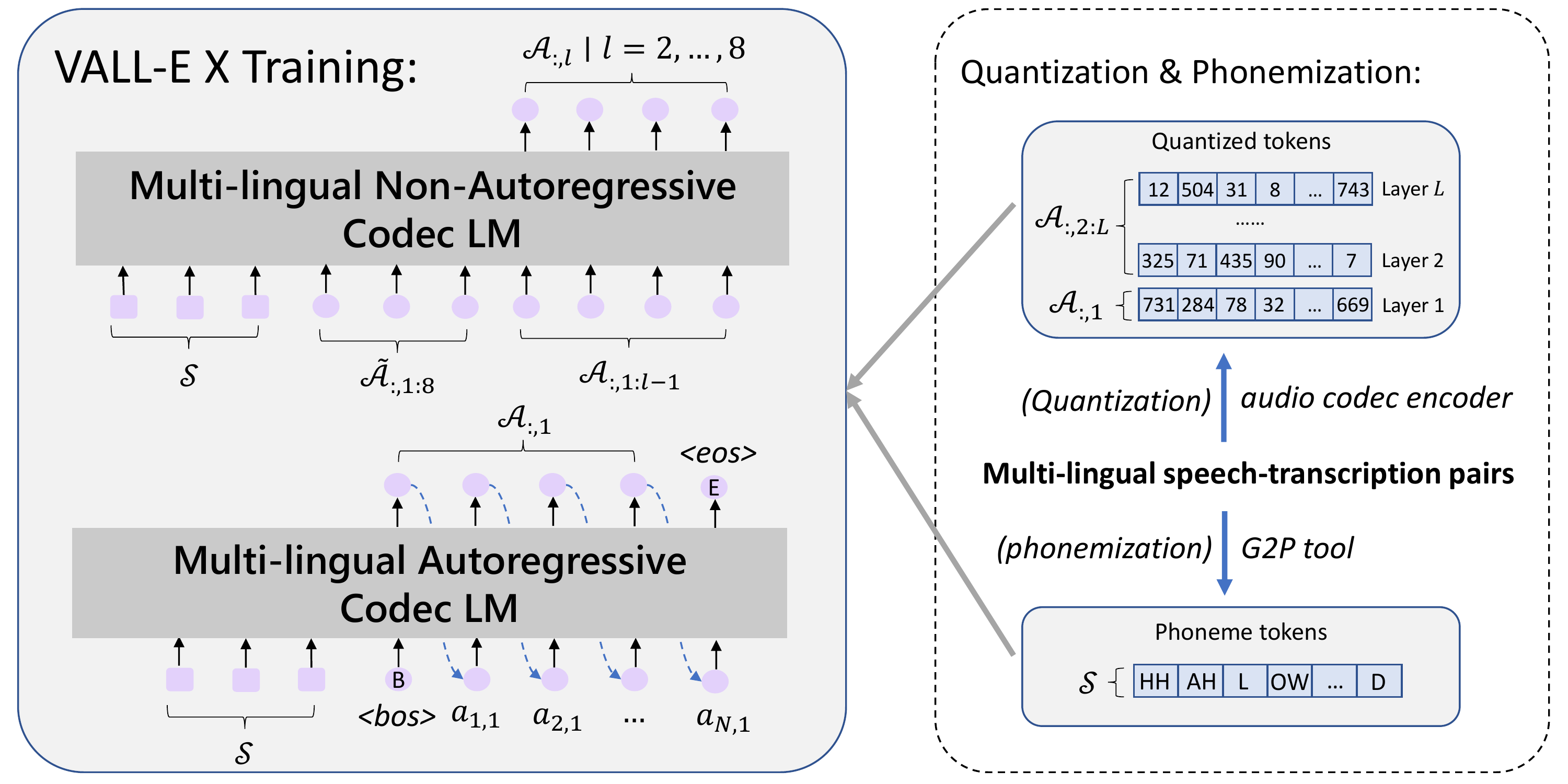}
  \caption{Training illustration of  the cross-lingual neural codec language model \our{}, consisting of a multi-lingual autoregressive codec LM ($\phi_{\mathrm{MAR}}$) and a multi-lingual non-autoregressive codec LM ($\phi_{\mathrm{MNAR}}$). 
  Multi-lingual acoustic tokens ($\mathcal{A}$) and phoneme sequences ($\mathcal{S}$) are converted from speech and transcription using an audio codec encoder and G2P tool, respectively.
  During training, we use paired $\mathcal{S}$ and $\mathcal{A}$ from different languages to optimize these two models. 
  }
  \label{fig_vallex_training}
\end{figure*}

\subsection{Model Framework}
\label{section_speechgen}

Inspired by VALL-E, the cross-lingual codec language model \our{} (denoted as $\phi$) leverages a multi-lingual autoregressive codec LM and a multi-lingual non-autoregressive codec LM to generate acoustic tokens at different granularities, as shown in the left part of Figure \ref{fig_vallex_training}.
We also adopt the neural codec model EnCodec \citep{defossez2022high} as the acoustic quantizer, which is an encoder-decoder model with $L$ quantization layers.
We choose $L=8$ in our experiments, for each layer it produces quantized codes of 1024 entries at 75Hz.

\paragraph{Multi-lingual Autoregressive Codec LM}
The multi-lingual autoregressive codec LM  $\phi_{\mathrm{MAR}}$ is a unidirectional Transformer decoder that autoregressively generates acoustic tokens based on the semantic tokens (phoneme sequence).
To make the sentence-level training efficient and accelerate the decoding during inference, similar to VALL-E, the cross-lingual autoregressive codec LM $\phi_{\mathrm{MAR}}$ is only used to predict the acoustic tokens from the first quantizer of EnCodec model.

Formally, based on paired speech-transcription data in any language, let $\mathcal{S}$ denote the transcribed phoneme sequence, and $\mathcal{A}_{:,1} \triangleq \{a_{i,1}|i=1,\ldots,N\}$ denotes the first-layer acoustic tokens extracted from the speech $\mathcal{X}$.
The decoder $\phi_{\mathrm{MAR}}$, modeling the concatenated sequence $\left \langle \mathcal{S},\mathcal{A}_{:,1}\right \rangle$, is trained to predict $\mathcal{A}_{:,1}$ autoregressively.
It is optimized by maximizing the log-likelihood, 
\begin{equation}
\mathcal{L}_{\mathrm{MAR}} =- \mathrm{log}~p_{\mathrm{AR}}\left(\mathcal{A}_{:,1}\mid\mathcal{S};\phi_{\mathrm{MAR}}\right) = - \mathrm{log}~ \prod_{i=1}^{N}p\left(a_{i,1}\mid\left \langle \mathcal{S},\mathcal{A}_{<i,1}\right \rangle;\phi_{\mathrm{MAR}} \right)
\end{equation}
where $\langle \rangle$ means sequence concatenation operation, and $p(.)$ is the softmax function.

\paragraph{Multi-lingual Non-Autoregressive Codec LM} 
Instead of the autoregressive generation pattern, multi-lingual non-autoregressive codec LM $\phi_{\mathrm{MNAR}}$ is a non-autoregressive Transformer language model aiming at iteratively generating the rest layers of acoustic tokens from the first layer.
It is prompted by the phoneme sequence of the current sentence ($\mathcal{S}$) and the acoustic token sequence of another sentence with the same speaker ($\mathcal{\tilde{A}}$).
Here $\mathcal{\tilde{A}}$ is taken from the previous sentence in the dataset where the adjusted sentences are usually segmented from the same paragraph. 
It is expected to have the same characteristics of voice (speaker, speed, background, etc) as the current sentence and is used as an additional reference for cloning the target voice.
Like VALL-E, for generating acoustic tokens of each layer $l\in [2,8]$, the embeddings of $l-1$ layers' acoustic tokens ($\mathcal{A}_{:,1:l-1}$) are summed up layerwise as input.
The learning objective for the $l$-layer acoustic tokens $\mathcal{A}_{:,l}$ can be calculated as
\begin{equation}
    \mathcal{L}_{\mathrm{MNAR}} = \sum_{l=2}^{8} \mathrm{log}~p_{\mathrm{NAR}}\left(\mathcal{A}_{:,l}\mid\left\langle\mathcal{S},\mathcal{\tilde{A}}_{:,1:8},\mathcal{A}_{:,1:l-1}\right\rangle;\phi_{\mathrm{MNAR}}\right) \label{eqn:ate_loss}
\end{equation}
where $\langle \rangle$ means the sequence concatenation.
$p_{\mathrm{NAR}}(.)$ computes the pointwise probabilities of $\mathcal{A}_{:,l}$.

\subsection{Multi-lingual Training} \label{ssec:bilingual_training}

In order to learn cross-lingual acoustic conversion information for cross-lingual TTS and speech-to-speech translation tasks, we take advantage of bilingual speech-transcription (ASR) corpus\footnote{Current version of \our{} is trained on the speech-transcription of two languages, we leave exploring more languages for future work.}, pairs of ($\mathcal{S}^{s}$, $\mathcal{A}^{s}$) and ($\mathcal{S}^{t}$, $\mathcal{A}^{t}$) to train our multi-lingual codec LMs $\phi_{\mathrm{MAR}}$  and $\phi_{\mathrm{MNAR}}$, where $s$ and $t$ represent two different (source and target) languages.

\paragraph{Language ID Module}
Following multi-lingual TTS, we leverage a language ID to guide the speech generation for specific languages in \our{}.
On the one hand, without language ID, \our{} may be confused to select suitable acoustic tokens for the specific language since it is trained with multi-lingual data.
On the other hand, some languages have very different characteristics, for example, Chinese is a tone language while English is a non-tone language, which increases the difficulty of adjusting the speaking style across languages.
Our experiments found that adding language information to the input of our multi-lingual autoregressive codec LM $\phi_{\mathrm{MAR}}$ is surprisingly effective in guiding the right speaking style and relieving the L2 accent problem, which will be introduced in Section \ref{analysis}.
Concretely, we embed language IDs into dense vectors and add them to the embeddings of acoustic tokens.

\subsection{Cross-Lingual Inference}
\label{subsec_cross_lingual_inference}

\begin{figure*}[!t]
  \centering
  \includegraphics[width=14cm]{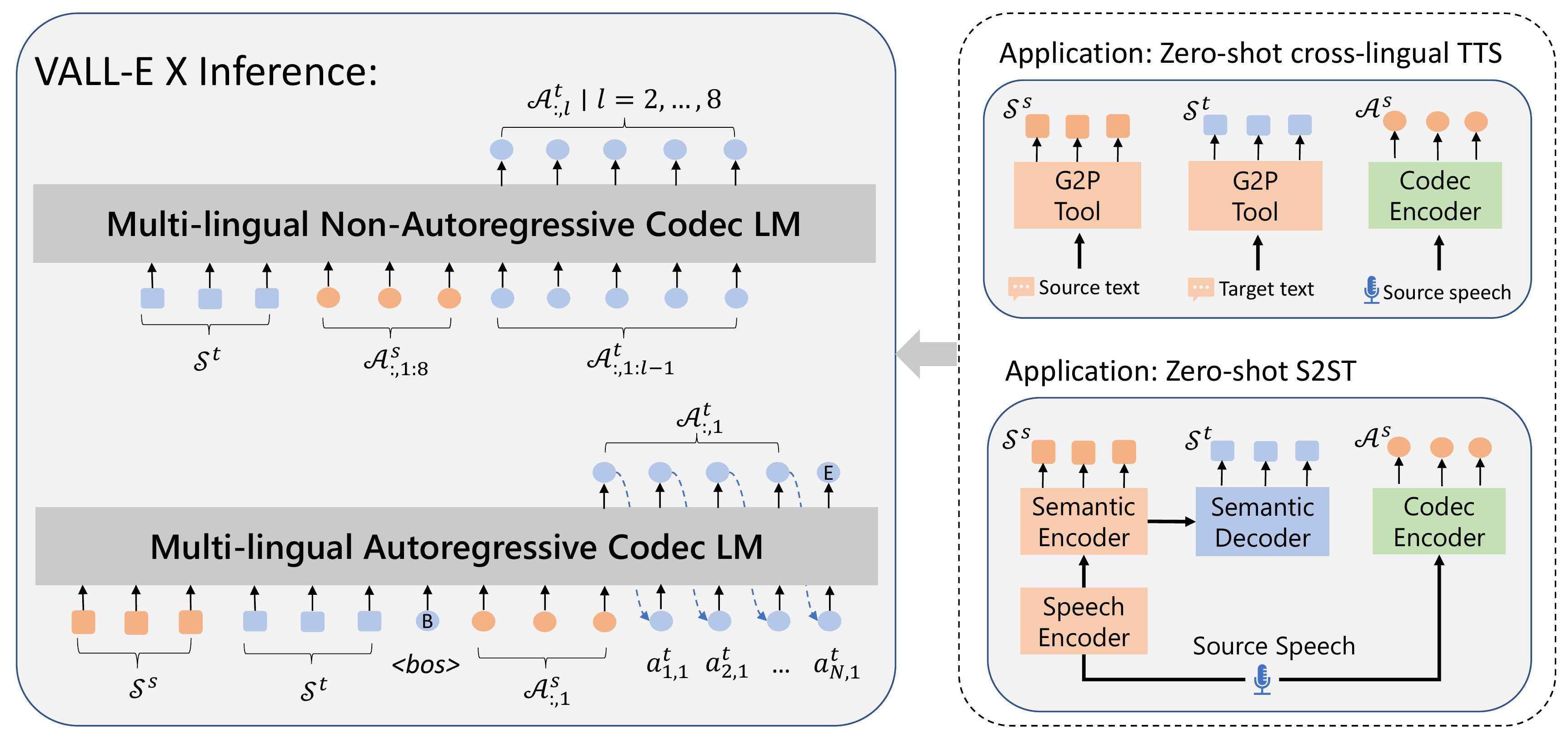}
  \caption{Inference illustration of  the cross-lingual neural codec language model \our{}, with two-stage decoding strategies.
  \our{} can support zero-shot cross-lingual TTS and zero-shot speech-to-speech translation tasks.
  }
  \label{fig_vallex_inference}
\end{figure*}

After training, \our{} can perform cross-lingual speech synthesis, as shown in Figure \ref{fig_vallex_inference}.
In detail, we first concatenate source phonemes $\mathcal{S}^{s}$ and target phonemes $\mathcal{S}^{t}$ as prompts, and take the first-layer source acoustic tokens $\mathcal{A}^{s}_{:,1}$ as the decoding prefix, condition on which the multi-lingual autoregressive codec LM $\phi_{\mathrm{MAR}}$ generates the first-layer target acoustic tokens $\mathcal{A}^{t}_{:,1}$,
\begin{equation}
\hat{a}^{t}_{i,1} \sim p_{\mathrm{AR}}\left(a^t_{i,1}\mid \left\langle \mathcal{S}^{s},\mathcal{S}^{t},\mathcal{A}^{s}_{:,1},\mathcal{A}^{t}_{<i,1}\right\rangle;\phi_{\mathrm{MAR}}\right), i=1,\ldots,
\label{equation_ar_infenrece}
\end{equation}
where $\sim$ means probability-based sampling.
The sampling is stopped until the \texttt{<end-of-sentence>} token is sampled.
As mentioned in Section \ref{ssec:bilingual_training}, language ID is used to control the speaking style of the final generated speech.
After obtaining the first-layer target acoustic tokens $\mathcal{A}^{t}_{:,1}$ from $\phi_{\mathrm{MAR}}$, multi-lingual non-autoregressive codec LM $\phi_{\mathrm{MNAR}}$ is used to predict the rest layers of acoustic tokens $\left\{\mathcal{A}^{t}_{:,l}\mid l=2,\ldots,8\right\}$ by greedy search, i.e., choosing the tokens with maximum probabilities,
\begin{equation}
\label{equation_nar_infenrece}
\mathcal{A}^t_{:,l} = \mathop{\mathrm{argmax}}\limits_{\mathcal{A}^t_{:,l}} p_{\mathrm{NAR}}\left(\mathcal{A}^t_{:,l}\mid \left\langle \mathcal{S}^{t},\mathcal{A}^{s}_{:,1:8},\mathcal{A}^t_{:,1:l-1}\right\rangle; \phi_{\mathrm{MNAR}} \right), l=2,\ldots,8.
\end{equation}
Finally, we use the decoder of EnCodec to synthesize the target speech from the complete target acoustic tokens $\mathcal{A}^{t}_{:,1:8}$.

\section{\our{} Application}

\our{} can be applied to various cross-lingual speech generation tasks. In this paper, we take zero-shot cross-lingual TTS and zero-shot speech-to-speech translation as two examples, as illustrated in Figure \ref{fig_vallex_inference}. 

\subsection{Zero-Shot Cross-Lingual TTS}
The proposed \our{} is naturally suitable for zero-shot cross-lingual TTS tasks.
Cross-lingual TTS tries to synthesize the target speech from text with a foreign speaker's voice.
Conventional methods mainly employ additional speaker and language networks to model the speaker and language information respectively, without zero-shot synthesis capability.
Thanks to the in-context learning capability of large language models, \our{} surprisingly shows the ability to perform zero-shot cross-lingual speech synthesis.
More specifically, given the source speech, source transcript, and target text, we first convert source speech into source acoustic token $\mathcal{A}^{s}$ using the encoder of neural codec model EnCodec, and convert source transcript and target text into source phonemes $\mathcal{S}^{s}$ and target phonemes $\mathcal{S}^{t}$ using G2P tool.
More specifically, as introduced in Section \ref{subsec_cross_lingual_inference}, we let $\mathcal{S}^{t}$ be the phonemes extracted from the target text, $\mathcal{S}^{s}$ and $\mathcal{A}^{s}$ be the phonemes and acoustic tokens extracted from the source speech. 
Then \our{} generates the full-layer target acoustic tokens, which are finally decompressed into the target speech by EnCodec decoder.

\subsection{Zero-Shot Speech-to-Speech Translation}

We can also apply our \our{} to zero-shot speech-to-speech translation tasks with additional speech recognition \& translation model, which is responsible for synchronously recognizing and translating the source speech to the source and target phoneme sequences.

\paragraph{Speech Recognition \& Translation Model}
We leverage the improved SpeechUT \citep{zhang2022speechut} as our speech recognition \& translation model, which is a unified-modal speech-unit-text pre-training framework using hidden units as the modality bridge between speech and text.
It supports various speech-to-text tasks, including both ASR and speech-to-text translation (ST).
Inspired by SpeechLM \citep{zhang2022speechlm} which explores different choices of units, we improve SpeechUT by replacing the clustering-based hidden units with phonemes.
Specifically, it consists of a speech encoder, a phoneme encoder, and a phoneme decoder.
All these components are pre-trained on ASR corpus (source speech $\mathcal{X}^\mathrm{{s}}$, source phoneme $\mathcal{S}^\mathrm{{s}}$) and MT corpus (source phoneme $\mathcal{S}^\mathrm{{s}}$, target phoneme $\mathcal{S}^\mathrm{{t}}$), where the phoneme sequences are converted from the text. 
Please see Appendix \ref{subsec_Appendix_speechut_td} for more pre-training details about this model.
After pre-training, the model is fine-tuned with ($\mathcal{X}^s$, $\mathcal{S}^s$, $\mathcal{S}^t$) triplet data derived from the ST corpus.
Specifically, we perform multi-task learning with the CTC \citep{graves2006ctc} loss added on the phoneme encoder predicting the source phonemes and the cross-entropy loss on the phoneme decoder predicting the target phonemes.

\paragraph{Inference}
Figure \ref{fig_vallex_inference} shows the inference process of speech-to-speech translation. Given a source speech $\mathcal{X}^{s}$, the speech recognition \& translation model first generates the source phonemes $\mathcal{S}^{s}$ from the semantic encoder and the target  phonemes $\mathcal{S}^{t}$ from the semantic decoder.
Besides, we use the EnCodec encoder to compress $\mathcal{X}^{s}$ into source acoustic tokens $\mathcal{A}^{s}$.
Then, we concatenate $\mathcal{S}^{s}$, $\mathcal{S}^{t}$, and $\mathcal{A}^{s}$, as the input of \our{}, to produce the acoustic token sequence for the target speech, as introduced in Section \ref{subsec_cross_lingual_inference}. The generated acoustic tokens are converted to the final target speech with the decoder of EnCodec.

\subsection{Evaluation} \label{ssec:evaluation}
The proposed model is verified using various evaluation criteria, including speaker similarity (ASV-Score), speech quality (ASR-WER), translation quality (ASR-BLEU), naturalness, and human evaluation.
Specifically, we measure speaker similarity between synthesized target speech and groud-truth target speech or source speech as an automatic speaker verification (ASV) task, where a WavLM \citep{chen2022wavlm} based ASV model is used to calculate the score.
To verify the quality of generated speech, we first utilize the ASR system from the released HuBERT-Large model \citep{hsu2021hubert} to recognize it into text.
For TTS, speech quality is measured by ASR-WER between the recognized text and the original target text.
For S2ST, speech quality is measured by ASR-BLEU between the recognized text and the provided translation text.
Finally, to better verify our proposed \our{} systems, we adopt the open-source NISQA\footnote{https://github.com/gabrielmittag/NISQA} \citep{mittag2021deep} (the NISQA-TTS model) to evaluate the naturalness of the synthetic speech and further perform the human evaluation with manual scoring on the generated speech, e.g., mean opinion score (MOS), comparative mean opinion score (CMOS) and similar mean opinion score (SMOS).

\section{Experiments}
\label{experiments}
We evaluate the proposed model on zero-shot cross-lingual TTS including English TTS prompted by Chinese speakers and Chinese TTS prompted by English speakers, and zero-shot S2ST including Chinese$\rightarrow$English and English$\rightarrow$Chinese directions.
We provide the synthesized audio samples on our demo page to better show the performance of \our{}.

\subsection{Dataset} \label{ssec:data}
Our \our{} is trained using bilingual speech-transcription (ASR) data.
The Chinese ASR data are from WenetSpeech \citep{Zhang2022WENETSPEECH} containing 10,000+ hours of multi-domain labeled speech.
The English ASR data are from LibriLight \citep{Kahn2020LibriLight} containing about 60,000 hours of unlabeled speech, whose speech data are collected from audiobooks.
We train a Kaldi\footnote{https://github.com/kaldi-asr/kaldi/tree/master/egs/librispeech} ASR model on the labeled Librispeech \citep{Panayotov2015Librispeech} dataset to generate the pseudo transcripts for the unlabeled LibriLight speech.

To train the speech recognition \& translation model for S2ST, we also use additional MT and ST data.
The MT data are from AI Challenger\footnote{https://challenger.ai/competition/translation}, OpenSubtitles2018\footnote{https://opus.nlpl.eu/OpenSubtitles2018.php} and WMT2020\footnote{https://www.statmt.org/wmt20/translation-task.html}, which contain about 13M, 10M, and 50M sentence pairs in conversion, drama\footnote{http://www.opensubtitles.org/}, and news domains, respectively.
The English$\rightarrow$Chinese ST data is from GigaST \citep{ye2022gigast}, which is created by translating the transcripts in GigaSpeech \citep{chen2021gigaspeech} using a strong machine translation system.
Similarly, we create the Chinese$\rightarrow$English ST data by translating the transcripts of WenetSpeech using an MT model trained by ourselves on the MT data mentioned above.

We evaluate zero-shot S2ST using the Effective Multilingual Interaction in Mobile Environments (EMIME) dataset \citep{wester2010emime}, which contains bilingual Chinese/English speech recorded by the same speakers.
There are 25 pairs of bilingual sentences recorded by 7 female and 7 male native Chinese speakers, thus the total number of test examples is 350.
Zero-shot cross-lingual TTS is evaluated using Librispeech \citep{Panayotov2015Librispeech} dev-clean set and EMIME dataset providing English and Chinese data, respectively.
We have two settings in the experiments: (1) Librispeech English TTS with EMIME Chinese speech as prompts; (2) EMIME Chinese TTS with Librispeech Engish speech as prompts.

\subsection{Experimental Setup}
\paragraph{Phonemization \& Quantization}
The right picture of Figure \ref{fig_vallex_training} illustrates the phonemization \& quantization processes for different languages.
All text data, including ASR transcripts and MT/ST translations, are converted by the lexicon provided in ASR datasets.
We use a unified phoneme set called BigCiDian\footnote{https://github.com/speechio/BigCiDian} for two languages which are based on International Phonetic Alphabet (IPA).
The ASR transcripts (or pseudo transcripts) are also converted by Kaldi force-alignment tools\footnote{https://github.com/kaldi-asr/kaldi/tree/master/} for additional alignment information used for the pre-training of speech recognition \& translation model.
The speech is quantized into discrete codec codes as the acoustic tokens using the neural audio codec model EnCodec\footnote{https://github.com/facebookresearch/encodec}, which employs residual vector quantization to iteratively quantize speech to a codebook according to the residual after quantization, resulting in multi-layer codebooks.

\paragraph{Model Architecture}
For the cross-lingual codec language models, $\phi_{\mathrm{MAR}}$  and $\phi_{\mathrm{MNAR}}$  are both 12-layer Transformer decoders with an attention dimension of 1024 and the FFN dimension of 4096.
The autoregression is implemented by attention masking in the $\phi_{\mathrm{MAR}}$  model.
Sinuous position embedding is separately computed for each prompt sequence in $\phi_{\mathrm{MAR}}$ and $\phi_{\mathrm{MNAR}}$ models.
Besides, the $\phi_{\mathrm{MNAR}}$  model uses individual layer normalization for generating each layer of acoustic tokens. We also introduce the model architecture of speech recognition \& translation for S2ST in Appendix \ref{subsec_Appendix_speechut_ma}. We call our cross-lingual TTS model and S2ST model as \textbf{\our{}} and \textbf{\our{} Trans} in the subsequent experiments, respectively.

\paragraph{Training Details}
We optimize each module of \our{} individually, including $\phi_{\mathrm{MAR}}$ and $\phi_{\mathrm{MNAR}}$.
For both modules, The maximum sentence length is set to 20 seconds, so we re-segment the LibriLight data to an average utterance duration of 12 seconds by detecting the consecutive silence phonemes.
Fortunately, the WenetSpeech data has already been segmented into short utterances.
The maximum learning rate is 5e-4 with warm-up steps of 8,000.
The models are trained on 32 V100 GPUs for 800k steps.
$\phi_{\mathrm{MAR}}$ is trained with the batch size of 120 seconds per GPU, which is 66 seconds for $\phi_{\mathrm{MNAR}}$ due to the memory constraint.
When optimizing $\phi_{\mathrm{MNAR}}$, instead of accumulating all layer's loss in Eqn. (\ref{eqn:ate_loss}), we randomly select one layer at each optimization step for efficiency.
For speech recognition \& translation model, the training details can be found in Appendix \ref{subsec_Appendix_speechut_td}.

\paragraph{Baselines}
We adopt YourTTS\footnote{https://github.com/Edresson/YourTTS} \citep{casanova2022yourtts} as our baseline for zero-shot cross-lingual TTS. YourTTS is a zero-shot multi-speaker TTS model for everyone, whose speaker information is based on speaker embedding extracted from a reference speech.
Since previous work shows that current end-to-end S2ST systems underperform cascaded S2ST systems \citep{jia2022translatotron,lee2021textless}, we also build an S2ST baseline which is cascaded by an ASR model, an MT model, and a multi-speaker YourTTS.
The source speech serves as the reference speech when synthesizing the target speech using YourTTS.
The ASR model is the released HuBERT model introduced in Section~\ref{ssec:evaluation}, and the MT model is a vanilla Transformer trained by ourselves on the MT data introduced in Section~\ref{ssec:data}.
Since YourTTS is built only for English, we don't get its performance for English$\rightarrow$Chinese translation direction.

\begin{table}[ht]
\caption{Zero-shot cross-lingual TTS evaluation for English TTS with Chinese speech as prompts and Chinese TTS with English speech as prompts, using automatic evaluation matrices, including ASV-Score (hypothesis vs. prompt), ASR-WER, and Naturalness.}
\centering
\begin{tabular}{l|ccc}
\toprule
 & ASV-Score     & ASR-WER  & Naturalness       \\
\midrule
\multicolumn{4}{l}{\textit{English TTS with Chinese as prompts}} \\
\midrule
Baseline (YourTTS)                  & 0.30$\pm$0.10          & 8.53  & 3.36  \\
\our{}                         & 0.36$\pm$0.11 & 4.07 & 3.54 \\
\midrule
\multicolumn{4}{l}{\textit{Chinese TTS with English as prompts}}  \\
\midrule
\our{}                         & 0.29$\pm$0.10 & 8.52 & 3.36  \\
\bottomrule
\end{tabular}
\label{Tab:x_tts}
\end{table}

\subsection{Zero-Shot Cross-Lingual TTS Evaluation}
We first select samples with a length between 4 and 10 seconds from LibriSpeech dev-clean set, resulting in 40 speakers and 1373 samples.
For English TTS, we randomly select one audio from EMIME set as the Chinese prompt for each target sentence in LibriSpeech dev-clean set.
For Chinese TTS, we use extra 149 Chinese text sentences provided by the EMIME set and repeat them to the total number of 1373 so that they can be prompted by the LibriSpeech audios one-by-one.
When synthesizing the target language speech, the whole sequence of the source language speech is used as the prompt.

\paragraph{Automatic Evaluation}
Table \ref{Tab:x_tts} summarizes the results of cross-lingual zero-shot TTS tasks, including English TTS prompted by Chinese speech and Chinese TTS prompted by English speech.
We measure the speaker similarity using the automatic speaker verification (ASV) model, ranging from -1 to +1 given two speech utterances. The larger the value, the more similar the speakers of the two utterances are.
The results show that: (1) for English TTS with Chinese as prompts, the speaker similarity between the hypothesis and prompts of \our{} is superior to that of the baseline (0.36 vs 0.30). (2) \our{} reduces the WER significantly from the baseline (from 8.53 to 4.07), demonstrating the effectiveness of our method. (3) \our{} has better speech naturalness than the baseline thanks to the large-scale training data and the large language model capacity.
The results of Chinese TTS with English prompts are also listed.

\begin{table}[h]
\begin{center}
\caption{Human evaluation for zero-shot cross-lingual TTS. SMOS means similarity MOS between generated speech and prompt, and CMOS means comparative MOS based on Baseline.
\label{exp:libri_spk_human} }
\begin{tabular}{lcc}
\toprule
& SMOS & CMOS (v.s. Baseline)   \\
\midrule
Baseline (YourTTS)  & 3.42$\pm$0.19 &  0.00 \\ 
\our{} & 4.00$\pm$0.20 & +0.24\\ 
 \bottomrule
\end{tabular}
\end{center}
\end{table}

\paragraph{Human Evaluation}

We further conduct the human evaluation on 50 randomly selected speech records for zero-shot cross-lingual English TTS with Chinese speech as prompts, including SMOS and CMOS. 
Note that SMOS ranges from 1 to 5 where the larger the value, the higher the voice similarity, and CMOS ranges from -3 to 3 where the positive number means the new system is better than the baseline. The results are listed in Table \ref{exp:libri_spk_human}. Baseline gets 3.42 SMOS scores between generated speech and prompts, while our \our{} achieves 4.00, which further demonstrates the model's superiority in keeping the speech characteristic in the cross-lingual setting.
Moreover, to directly compare the speech synthesis quality between the proposed \our{} and baseline, we calculate the CMOS score between them evaluated by native speakers on the 50 sentences. The last column of Table \ref{exp:libri_spk_human} shows that \our{} obtains the gain of +0.24 CMOS scores than the baseline.

\subsection{Zero-Shot S2ST Evaluation}

S2ST is evaluated on bidirectional Chinese$\leftrightarrow$English data of EMIME dataset, measured by speaker similarity, translation quality, speech naturalness, and human evaluation.

\paragraph{Speaker Similarity}  We first evaluate whether the speaker's voice is preserved in the generated target speech using speaker similarity (ASV-Score), whose results are listed in Table \ref{Tab:results}.
Because the EMIME test set has paired speech utterances with Chinese and English, we are able to calculate the ASV score among the generated speech (hyp), the source speech (src), as well as the target speech (tgt), resulting in 3 settings (tgt vs. src, hyp vs. src, and hyp vs. tgt).
From Table \ref{Tab:results} we can find that:
(1) For Chinese$\rightarrow$English, the ASV score of \our{} Trans significantly outperforms that of the conventional speaker embedding based S2ST system (Baseline), demonstrating the superiority of our model in terms of maintaining the source speaker's voice.
(2) The ASV score has similar values when the generated speech (hyp) is compared with the source speech (src) and the target speech (tgt), and it is far away from the upper bound (tgt vs. src) for the English$\rightarrow$Chinese direction, which suggests that the cross-lingual voice transferability still has the improvement space.
(3) When directly generating speech from the ground-truth (oracle) text which degrades into cross-lingual TTS, the ASV score does not increase notably, indicating that voice transferability is less affected by the quality of translation.

\paragraph{Translation Quality} Table \ref{Tab:results} also shows the translation performance of \our{} Trans.
Note that ASR-BLEU with oracle target text as the input of \our{} can be seen as the upper bound when translations are exactly correct.
With oracle target text as input, \our{} Trans can achieve the performance of about 84$\sim$87 BLEU scores, which also reflects the high performance of our neural codec language model.
For Chinese$\rightarrow$English, \our{} Trans achieves higher BLEU over the baseline (30.66 vs. 27.49), demonstrating the end-to-end speech-to-phoneme translation is more effective against the conventional cascaded speech-to-text translation when applying to S2ST task.

\paragraph{Speech Naturalness}
We also evaluate the Naturalness with the open-source NISQA \citep{mittag2021deep} for S2ST outputs.
As shown in the last column of Table \ref{Tab:results}, compared to the baseline, \our{} Trans achieves a better naturalness score (3.54 vs. 3.44), which shows that \our{} can generate more natural target language speech than the baseline.

\begin{table}[t]
\caption{S2ST performance on EMIME dataset for Chinese$\leftrightarrow$English directions. Baseline is a cascaded S2ST system based on speaker embedding. Automatic evaluation matrices include ASV-Score, ASR-BLEU, and Naturalness.}
\centering
\small
\begin{tabular}{l|ccccc}
\toprule
 & \multicolumn{3}{c}{ASV-Score}         & \multirow{2}{*}{ASR-BLEU}     & \multirow{2}{*}{Naturalness} \\
 & tgt vs. src & hyp vs. src & hyp vs. tgt          &                               &   \\
\midrule
\multicolumn{6}{l}{\textit{Chinese$\rightarrow$English S2ST}} \\
\midrule
Baseline (S2ST)                   & \multirow{4}{*}{0.58$\pm$0.09}    & 0.28$\pm$0.10 & 0.27$\pm$0.12 & 27.49 & 3.44 \\
\quad - w/ oracle target text     &                                   & 0.28$\pm$0.10 & 0.29$\pm$0.11 & 80.30 & 3.43 \\
\our{} Trans                 &                                   & 0.37$\pm$0.10 & 0.37$\pm$0.11 & 30.66 & 3.54 \\
\quad - w/ oracle target text     &                                   & 0.39$\pm$0.10 & 0.38$\pm$0.10 & 86.78 & 3.54 \\
\midrule
\multicolumn{6}{l}{\textit{English$\rightarrow$Chinese S2ST}} \\
\midrule

\our{} Trans                 & \multirow{2}{*}{0.58$\pm$0.09}    & 0.48$\pm$0.11 & 0.53$\pm$0.11 & 34.45 & 3.41 \\
\quad - w/ oracle target text     &                                   & 0.47$\pm$0.12 & 0.55$\pm$0.11 & 84.00 & 3.42 \\
\bottomrule
\end{tabular}
\label{Tab:results}
\end{table}

\paragraph{Human Evaluation}

We randomly sample 56 translation pairs\footnote{There are 14 speakers in the bilingual Chinese/English dataset, and 4 sentence pairs are chosen for each speaker, resulting in 56 translation pairs in total.} to perform a human evaluation using SMOS and MOS matrics for both  Chinese$\rightarrow$English and English$\rightarrow$Chinese directions.
Table \ref{Tab:subjection_evaluation} lists the results of \our{} Trans as well as the Chinese$\rightarrow$English baseline.
We use MOS (from 1 to 5 scores) instead of CMOS because the translated content may be different among models, which is not suitable for CMOS evaluation.
For speaker similarity evaluation, \our{} Trans outperforms the baseline with 1.06 SMOS scores (4.12 vs. 3.06), demonstrating its superior ability to model speaker property of the proposed \our{}.
Note that this value still can be improved since it is still far from the SMOS between the source speech prompt and ground truth (4.91).
For speech quality, our \our{} slightly outperforms the baseline in Chinese$\rightarrow$English S2ST in terms of MOS score (3.87 vs. 3.81).

\begin{table}[h]
\caption{Subjection evaluation with SMOS and MOS scores on bidirectional Chinese$\leftrightarrow$English S2ST tasks. SMOS is measured by comparing with the ground-truth target speech. English$\rightarrow$Chinese S2ST baseline is not reported since it is not supported by the released YourTTS.}
\centering
\begin{tabular}{l|cc|cc}
\toprule
 & \multicolumn{2}{c|}{Chinese$\rightarrow$English} & \multicolumn{2}{c}{English$\rightarrow$Chinese} \\
 & SMOS & MOS  &  SMOS  & MOS \\
 \hline
Baseline (S2ST) &  3.06$\pm$0.14  & 3.81$\pm$0.19   & -  & -  \\
\our{} Trans &  4.12$\pm$0.13 &  3.87$\pm$0.21 & 3.94$\pm$0.15 & 3.48$\pm$0.13 \\
Source speech prompt & 4.91$\pm$0.05    & -  & 4.64$\pm$0.06 & -  \\
Oracle target speech & -    & 3.92$\pm$0.17  & - & 3.88$\pm$0.13  \\
\bottomrule
\end{tabular}

\label{Tab:subjection_evaluation}
\end{table}

\subsection{Analysis} \label{analysis}
In this section, we first analyze the effect of language ID, then explore the foreign accent problems, and qualitatively investigate the ability to maintain voice emotion and synthesize code-switch speech of our proposed model. 

\paragraph{Effect of Language ID}

Our \our{} is trained with multi-lingual ASR data, which might increase the modeling difficulty for each specific language. 
We address it by adding language IDs to guide speech synthesis in the autoregressive language codec model.
Here, we verify the effectiveness by removing the language ID (LID) or adding the wrong LID (i.e. the source LID).
The ASV-Score and ASR-BLEU are reported in Table \ref{Tab:lid_results}. Without LID or with the wrong language ID, the translation quality decreases, while the speaker similarity between the hypothesis and source speech increases. These results demonstrate the importance of language ID for the accuracy of the content.
It also indicates that target LID reduces the transfer of information, which means the model without LID or with source LID will better maintain the sound of the original speaker.

\begin{table}[ht]
\caption{Evaluation for the effect of language ID on  Chinese$\leftrightarrow$English EMIME dataset. ASV-Score is computed between synthesized speech and source prompt speech. The last column lists the subjection evaluation score of the foreign accent (from 1 to 5 scores).}
\centering
\small
\begin{tabular}{l|cc|c}
\toprule
 & ASV-Score (vs. src)     & ASR-BLEU  & Accent Score \\
\midrule
\multicolumn{4}{l}{\textit{Chinese$\rightarrow$English S2ST}} \\
\midrule
\our{} Trans                         & 0.37$\pm$0.10 &  30.66 & 4.10 \\
\quad w/o Language ID                & 0.41$\pm$0.10 & 29.04 & 2.98 \\
\quad w/ wrong Language ID          & 0.41$\pm$0.10 &  29.07 & 2.55 \\
\midrule
\multicolumn{4}{l}{\textit{English$\rightarrow$Chinese S2ST}} \\
\midrule
\our{} Trans                         & 0.48$\pm$0.11 &  34.45 & 4.03 \\
\quad w/o Language ID               & 0.49$\pm$0.11 &  30.86 & 2.35 \\
\quad w/ wrong Language ID          & 0.50$\pm$0.11 &  29.70 & 2.25 \\
\bottomrule
\end{tabular}
\label{Tab:lid_results}
\end{table}

\paragraph{Foreign Accent Control}

L2 (second-language, or foreign) accent problem, the synthesized speech sounds like the accents of a foreigner, has arisen in cross-lingual TTS systems \citep{zhang2019learning,lee2022empirical}.
Automatic Evaluation has shown that adding LID can boost speech quality.
Besides, we conduct a subjection evaluation to label foreign accents from 1 to 5 on randomly selected 20 synthesized speech for both English and Chinese, where each sample is measured with a score from 1 to 5 denoting high-status foreign speakers, low-status foreign speakers, middle-status speakers, low-status native speakers, and high-status native speakers, respectively.
As summarized in the last column of Table \ref{Tab:lid_results}, we observed that our \our{} can control the accent for the target speech by LID modules.
For example, in English$\rightarrow$Chinese, \our{} Trans with right LID and without LID get the score of 4.03 and 2.35, respectively.
This indicates that by using correct LID embedding, \our{} Trans is able to alleviate the foreign accent problem.
Please also see the demo for audio examples of \our{} Trans with or without language ID.

\paragraph{Voice Emotion Maintenance}

Generating the speech with a specific emotion is a difficult task for speech synthesis because conventional TTS methods require TTS data with emotion labels to train \citep{um2020emotional}. Moreover, it is more tempting to reserve the source speaker's emotion in generated target speech for the S2ST task, which is not explored in previous S2ST work.
In these experiments, we adopt the source prompts from the emotional voices dataset EmoV-DB \citep{um2020emotional} as inputs of \our{} Trans to generate the translated target speech, whose samples are listed on our demo page.
We found that the proposed \our{} can maintain emotional consistency to a certain extent between the source prompt and the synthesized speech. The underlying reasons are (1) our \our{} is trained with large-scale multi-lingual multi-speaker speech-transcription data, which contains various emotional speech records, and (2) the strong in-context learning ability of \our{}, like GPT-3 \citep{brown2020language}, promotes the generated speech to reserve the characteristic of the source prompt.

\paragraph{Code-Switch Speech Synthesis}

It is a common phenomenon to use code-switch utterances in bilingual or multi-lingual communities \citep{cao2020code,zhao2020towards,manghat2022normalization}. Code-switch speech synthesis aims to produce a fluent and  consistent voice for code-switch text. Although our proposed \our{} is trained on multiple monolingual speech data, without special optimization for code-switch setting, \our{} provides a promising solution to code-switch speech synthesis. We put the code-switch samples on our demo page, 
demonstrating that due to its strong in-context learning ability, \our{} can synthesize fluent code-switch speech with a consistent voice.

\section{Conclusion}
\label{conclusion}

In this work, we propose \our{}, a cross-lingual neural codec language model, which can retrain the source language speaker's voice in the generated target language speech.
\our{} is free of the requirement for cross-lingual paired data from the same speakers.
By training on large-scale multi-lingual multi-speaker speech-transcription data, the proposed \our{} demonstrates strong in-context learning capabilities and can support zero-shot cross-lingual text-to-speech and zero-shot voice-retentive speech-to-speech translation tasks.
For future work, we plan to expand this method with more data and more languages.


\bibliography{neurips_2022}
\bibliographystyle{plainnat}

\appendix

\section{Appendix}
\label{sec_Appendix}

\subsection{Speech Recognition \& Translation Model}
\label{subsec_Appendix_speechut}

\subsubsection{Model Pre-training}
\label{subsec_Appendix_speechut_pt}

Specifically, speech recognition \& translation model consists of a speech encoder ($\theta_{enc1}$), a semantic encoder ($\theta_{enc2}$), and a semantic decoder ($\theta_{dec}$).
Given a speech waveform $\mathcal{X}^s$ and the corresponding phonemes $\mathcal{S}^s \triangleq \{s^s_i|i=1,\ldots,N\}$ where $N$ is the sequence length, the speech-side pre-training objective is to predict the phonemes from the top of the speech encoder and semantic encoder, formalized as
\begin{equation}
\mathcal{L}_\mathrm{speech} =- \sum_{i\in \mathcal{M}}\left(\mathrm{log}~p\left(s^s_i|\mathcal{X}^s; \theta_{enc1}\right)+\mathrm{log}~p\left(s^s_i|\mathcal{X}^s;\theta_{enc1}, \theta_{enc2} \right)\right)
\end{equation}
where $\mathcal{M}$ is a set of masked positions, and the $p(.)$ is parameterized as the same way with original SpeechUT. 
Then, given bilingual phoneme sequences, $\mathcal{S}^s$ and $\mathcal{S}^t$, the text-side pre-training objective is to perform sequence-to-sequence translation autoregressively, formalized as
\begin{equation}
\mathcal{L}_\mathrm{text} =- \sum_{i=1}^{|\mathcal{S}^t|}\mathrm{log}~p\left(s^t_i|\mathcal{S}^t_{<i},\mathcal{S}^s;\theta_{enc2}, \theta_{dec} \right)
\end{equation}
In this way, each of the three components can be pre-trained with one or two learning objectives.
The final pre-training objective is
$\mathcal{L}_\mathrm{pt} = \mathcal{L}_\mathrm{speech} + \mathcal{L}_\mathrm{text}$. 


\subsubsection{Model Architecture}
\label{subsec_Appendix_speechut_ma}
For the speech recognition \& translation model, we leverage the Base architecture of the SpeechUT model, where all encoder/decoders consist of 6 Transformer layers with relative position bias \citep{shaw-etal-2018-self}.
The FFN dimension is 3072 and the attention dimension is 768.
Besides, a speech pre-net is equipped before the speech encoder, which contains several 1-D convolutional layers with 512 channels and kernel sizes of [10,3,3,3,3,2,2].
It can downsample the speech waveform by 320 and convert it to fix-dimensional embeddings.

\subsubsection{Training Details}
\label{subsec_Appendix_speechut_td}

The speech recognition \& translation model is pre-trained following the hyper-parameter setting of \citet{zhang2022speechut}.
The speech mask probability is 8\% and the mask length is 10.
The embedding mixing mechanism of the original SpeechUT is also performed.
The batch sizes of speech and phonemes on each GPU are 1,400,000 (87.5 seconds) and 3,000, respectively.
The maximum learning rate is 5e-4 with warm-up steps of 32,000.
The model is pre-trained on 32 V100 GPUs for 400K steps.
After pre-training, we perform ASR/ST joint fine-tuning, where the transcription phonemes are predicted on the top of the semantic encoder through a nonlinear CTC layer, and the translation phonemes are predicted through the semantic decoder.
the transcription phonemes are reduced by removing the repetitive phonemes.
During fine-tuning, we empirically set the weight of the CTC loss to 0.2.
The models are tuned on 32 GPUs with a batch size of 2,000,000 (125 seconds) per GPU for 200K steps.

\end{document}